\documentclass{article}
\usepackage{spconf,amsmath,graphicx}
\usepackage{multirow}
\usepackage{xspace}
\usepackage{url}
\usepackage{xspace}
\usepackage{hyperref}
\usepackage{cleveref}
\usepackage{color}
\usepackage{graphicx}
\usepackage{subcaption}

\usepackage{booktabs}
\usepackage{multirow}
\usepackage{mathtools}
\usepackage{amsmath,amssymb}
\usepackage{latexsym}
\usepackage{cite}
\usepackage{adjustbox}
\usepackage{wrapfig}
\usepackage{tikz,pgfplots}
\pgfplotsset{compat=newest}
\usetikzlibrary{plotmarks, shapes, arrows, positioning, shadows, trees}

\newcommand\BLEU{\textsc{Bleu}\xspace}
\newcommand\TER{\textsc{Ter}\xspace}

\newcommand\WER{\textsc{Wer}\xspace}

\newlength\figureheight 
	\newlength\figurewidth

\makeatletter

\renewcommand{\section}{\@startsection
 {section}%
 {1}%
 {}%
 {-0.7\baselineskip}%
 {0.3\baselineskip}%
 {}}%

\renewcommand{\subsection}{\@startsection
 {subsection}%
 {2}%
 {}%
 {-0.7\baselineskip}%
 {0.3\baselineskip}%
 {}}%

\renewcommand{\subsubsection}{\@startsection
 {subsubsection}%
 {3}%
 {}%
 {-0.7\baselineskip}%
 {0.3\baselineskip}%
 {}}%

\g@addto@macro\normalsize{%
 \setlength\abovedisplayskip{4pt plus 2pt minus 2pt}
 \setlength\belowdisplayskip{4pt plus 2pt minus 2pt}
 \setlength\abovedisplayshortskip{4pt plus 2pt minus 2pt}
 \setlength\belowdisplayshortskip{4pt plus 2pt minus 2pt}
}

\makeatother

\title{A Comparative Study on End-to-End Speech to Text Translation}
\name{Parnia Bahar$^{1,2}$, Tobias Bieschke$^1$, and Hermann Ney$^{1,2}$}
\address{
$^1$Human Language Technology and Pattern Recognition Group, Computer Science Department \\
RWTH Aachen University, 52074 Aachen, Germany, $^2$AppTek GmbH, 52062 Aachen, Germany \\
\texttt{\{bahar, ney\}@cs.rwth-aachen.de, tobias.bieschke@rwth-aachen.de}
}

\begin{document}
%\ninept
%
\maketitle
\begin{abstract}
Recent advances in deep learning show that end-to-end speech to text translation model is a promising approach to direct the speech translation field. 
In this work, we provide an overview of different end-to-end architectures,
as well as the usage of an auxiliary connectionist temporal classification (CTC) loss for better convergence.
We also investigate on pre-training variants such as initializing different components of a model using pretrained models, and their impact on the final performance, which gives boosts up to 4\% in \BLEU and 5\% in \TER. 
Our experiments are performed on 270h IWSLT TED-talks En$\to$De, and 100h LibriSpeech Audio-books En$\to$Fr.
We also show improvements over the current end-to-end state-of-the-art systems on both tasks.
\end{abstract}
\section{Introduction}
\label{sec:intro}
%Deep neural network (DNN) yields state-of-the-art performance in sequence modeling for many tasks if a moderate amount of training data is provided. 
The success of deep neural network (DNN) in both machine translation (MT) \cite{bahdanau_15_attention, sutskever_14_seq2seq, luong_15_attention, chen_18_best2worlds, bahar_2018_2d_mt} and automatic speech recognition (ASR) \cite{bahdanau_2016_asr, chan_2016_las, chorowski_2015_attention_asr, zeyer_2018_att_asr, bahar_2019_2d_asr} has inspired the work of end-to-end speech to text translation (ST) systems. 
The traditional ST methods are based on a consecutive cascaded pipeline of ASR and MT systems.
% and thus require transcribed source audio to train ASR and parallel text to train MT models. 
In contrast, 
the end-to-end stand-alone model \cite{berard_2016_proof, goldwater_2017_noasr, weiss_2017_directly, gangi-etal-2019-enhancing} translates speech in source language directly into target language text.
The end-to-end model has advantages over the cascaded pipeline,
 %Firstly, the end-to-end models avoid accumulating the error between the two systems. Secondly, they have lower latency in inference, and thirdly they require lower computational power in total.
however, its training requires a moderate amount of paired speech-to-text data which is not easy to acquire.
Therefore, recently some techniques such as multi-task learning \cite{weiss_2017_directly, anastasopoulos_2018_tied_multitask, Sperber_19_attention_passing, st_kd_interspeech2019},
pre-training different components of the model \cite{berard_2018_librispeech, bansal_2019_pretraining_asr, bansal_2018_low_resource} and generating synthetic data \cite{jia_2019_synthatic} have been proposed to mitigate the lack of ST parallel training data. These methods aim to use weakly supervised data, i.e. speech-to-transcription or text-to-translation pairs in addition to fully supervised data, i.e. speech-to-translation pairs. As reported in the literature, all of these methods give a boost up to some degree, however, each has its own problems. 

Multi-task learning has become a significant method that aims at improving
the generalization performance of a task using other related tasks \cite{luong_2015_multitask}. For this technique, we need to compromise between multiple tasks and the parameters are updated independently for each task, which might lead to a sub-optimal solution for the entire multi-task optimization problem. 
The pre-training methods and synthesis systems rely on given previously trained models. The component of the ST model can be trained using only an ASR model \cite{bansal_2019_pretraining_asr} or both an ASR and an MT model \cite{bansal_2018_low_resource}. Similarly, the synthetic techniques depend on a pre-trained MT or a text-to-speech (TTS) synthesis model. Both cases require more effort to build the models and thus more computational power. Furthermore, both scenarios rely on paired speech-to-transcription and/or text-to-translation data. In contrast, an unsupervised ST system uses only independent monolingual corpora of speech and text, though, its performance is behind the aforementioned approaches \cite{chung_2019_unsupervised_st}.

There are many aspects to be considered for training end-to-end models which we are exploring in this work. 
% , such as different network topologies, the use of an auxiliary CTC loss and various pretraining strategies. Our experiments show that there is a huge variance in translation performance depending on different aspects.
% In this paper, we aim to shed light on the following questions: 
% (1) which network architecture leads to the best performance?
% (2) 
Our contribution is as follows:
We review and compare different end-to-end methods for building an ST system and we analyze their effectiveness.
We confirm that multi-task learning is still beneficial up to some extent.
We also adopt the idea of connectionist temporal classification (CTC) auxiliary loss from ASR. Although the first intuition is that the CTC module can not cooperate with the ST task since the translation does not necessarily require the monotonic alignment of the speech and target sequences, surprisingly we show that including this loss helps to gain better performance. 
To facilitate the model with weakly supervised data, we study an extensive quantitative comparison of various possible pre-training schemes which to our best of knowledge, no such comparison for ST exists yet. 
We discover that directly coupling the pre-trained encoder and decoder hurts the performance and present a solution to incorporate them in a better fashion. 
We demonstrate our empirical results on two ST tasks, IWSLT TED-talks En$\to$De, and LibriSpeech Audio-books En$\to$Fr.
% and show an improvement over the end-to-end state-of-the-art systems.

\begin{figure*}[h]
    \centering
    \begin{subfigure}[b]{0.25\textwidth}
  \centerline{\includegraphics[width=3.5cm,height=5cm,keepaspectratio]{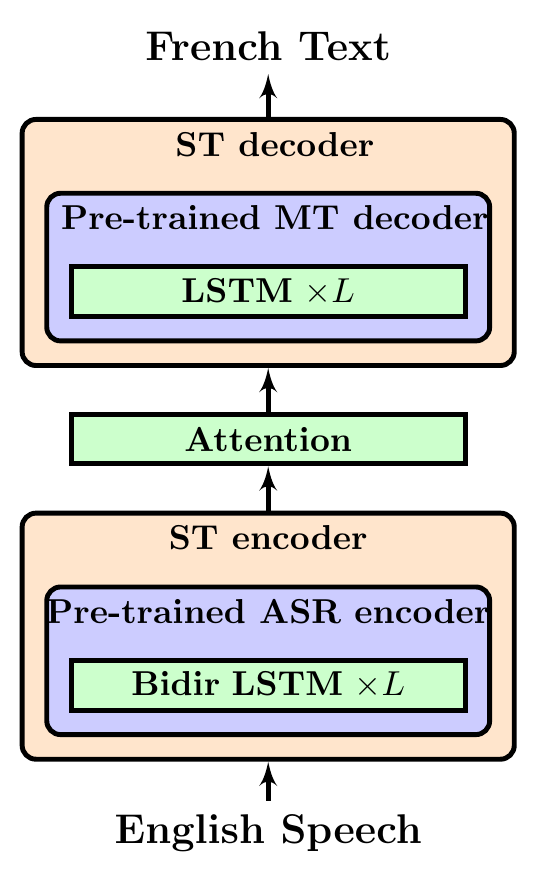}}
        \caption{direct}
        \label{fig:direct}
    \end{subfigure}
    ~\hfill %, \qquad, \hfill etc. 
    \begin{subfigure}[b]{0.35\textwidth}
  \centerline{\includegraphics[width=6.5cm,height=6.5cm,keepaspectratio]{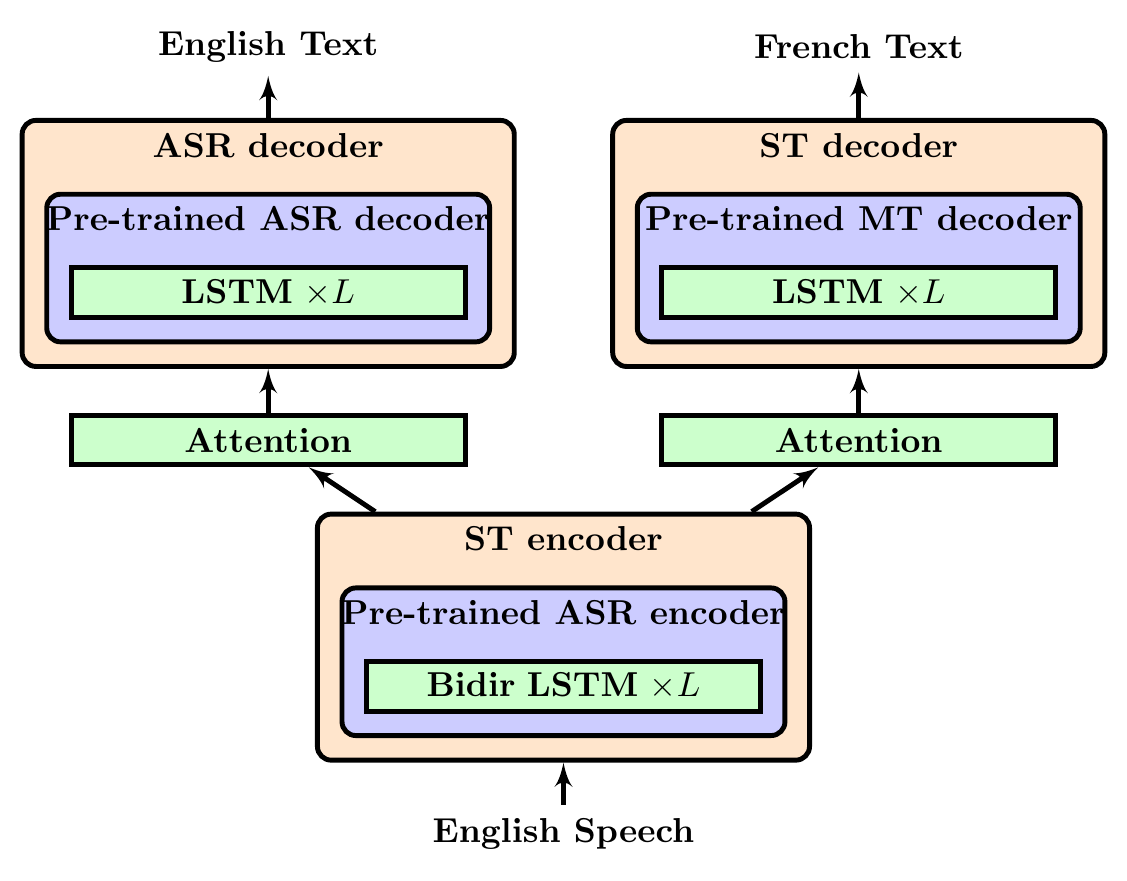}}
        \caption{multitask-one2many}
        \label{fig:multitask-one2many}
    \end{subfigure}
    ~\hfill
    \begin{subfigure}[b]{0.35\textwidth}
  \centerline{\includegraphics[width=6.5cm,height=6.5cm,keepaspectratio]{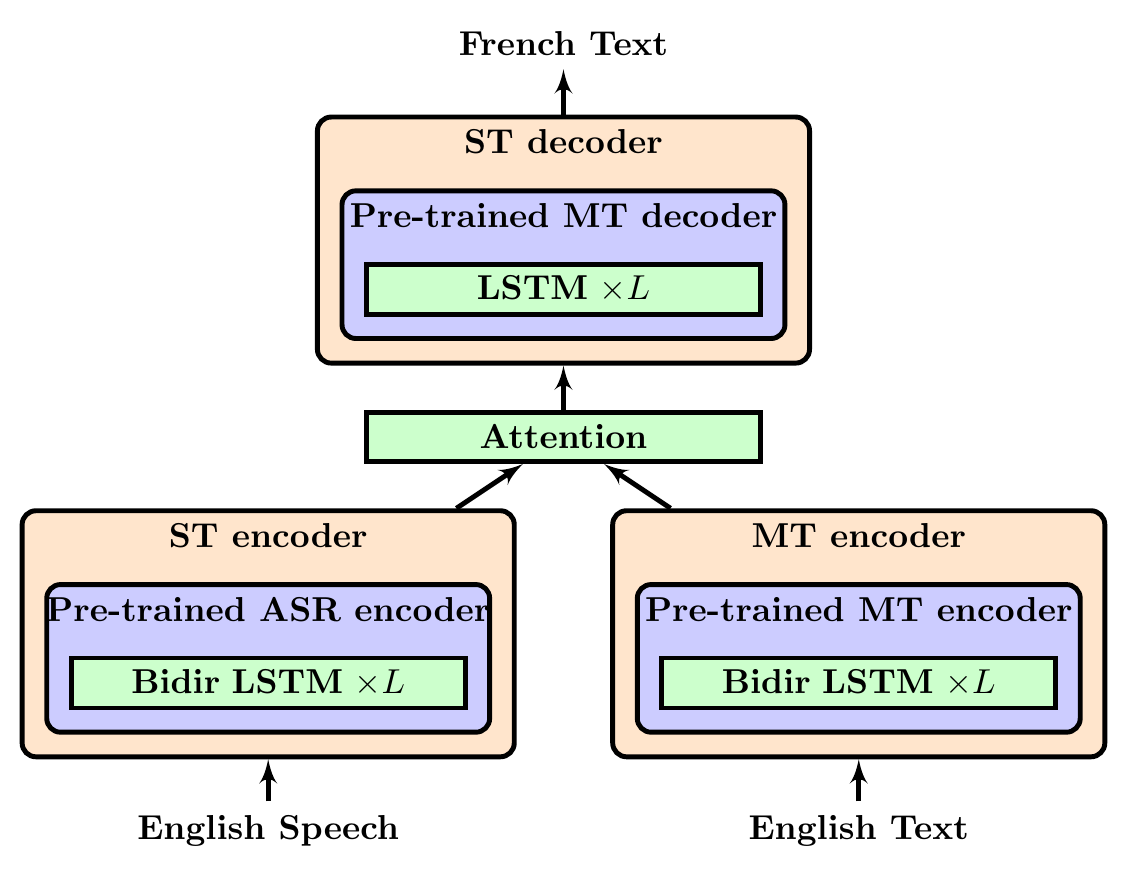}}
        \caption{multitask-many2one}
        \label{fig: multitask-many2one}
    \end{subfigure}
    
   \begin{subfigure}[b]{0.45\textwidth}
  \centerline{\includegraphics[width=6.8cm,height=6.8cm,keepaspectratio]{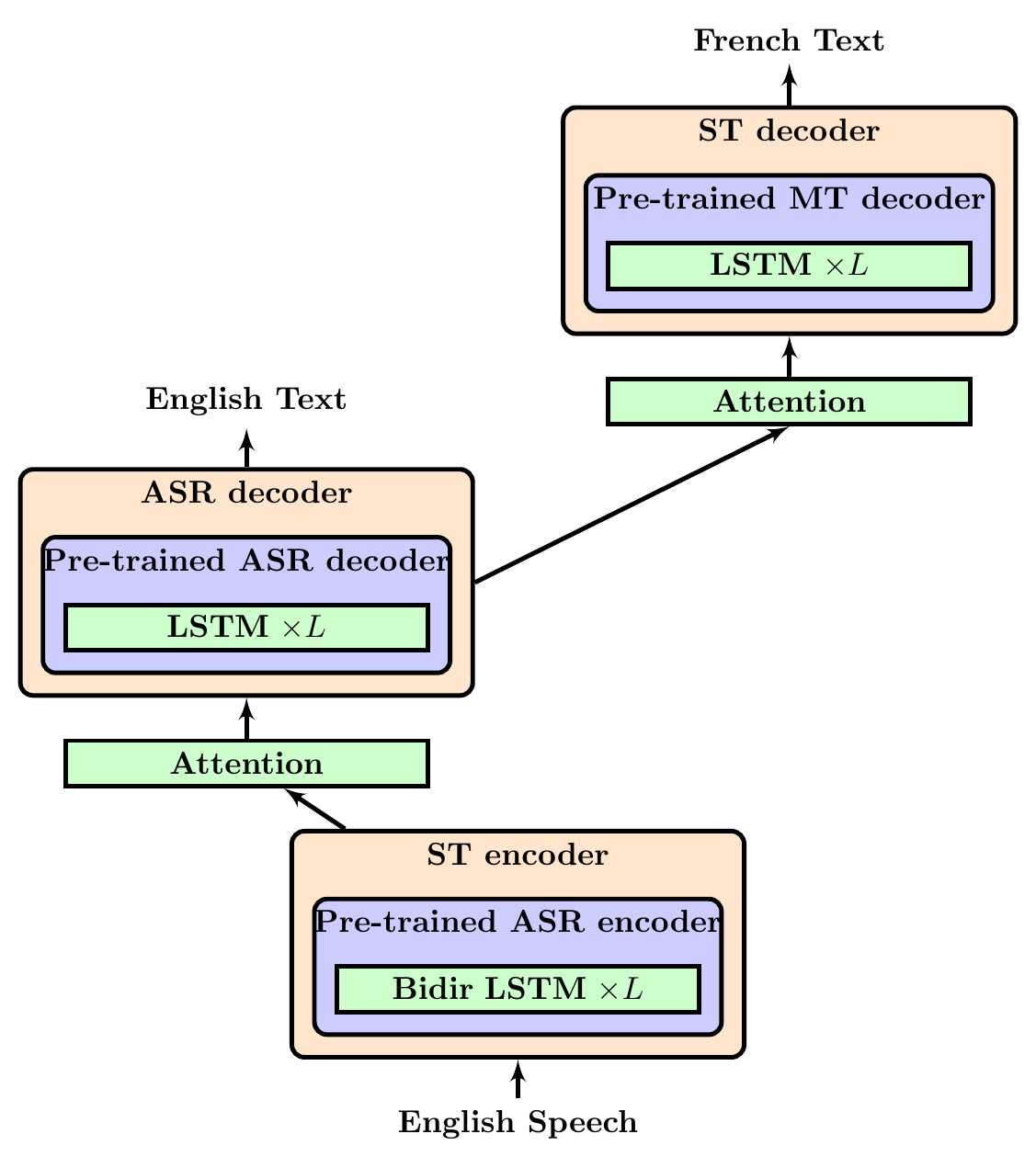}}
        \caption{multitask-tied cascade}
        \label{fig:multitask- tied cascaded}
    \end{subfigure}
    ~\hfill
    \begin{subfigure}[b]{0.45\textwidth}
  \centerline{\includegraphics[width=6.8cm,height=6.8cm,keepaspectratio]{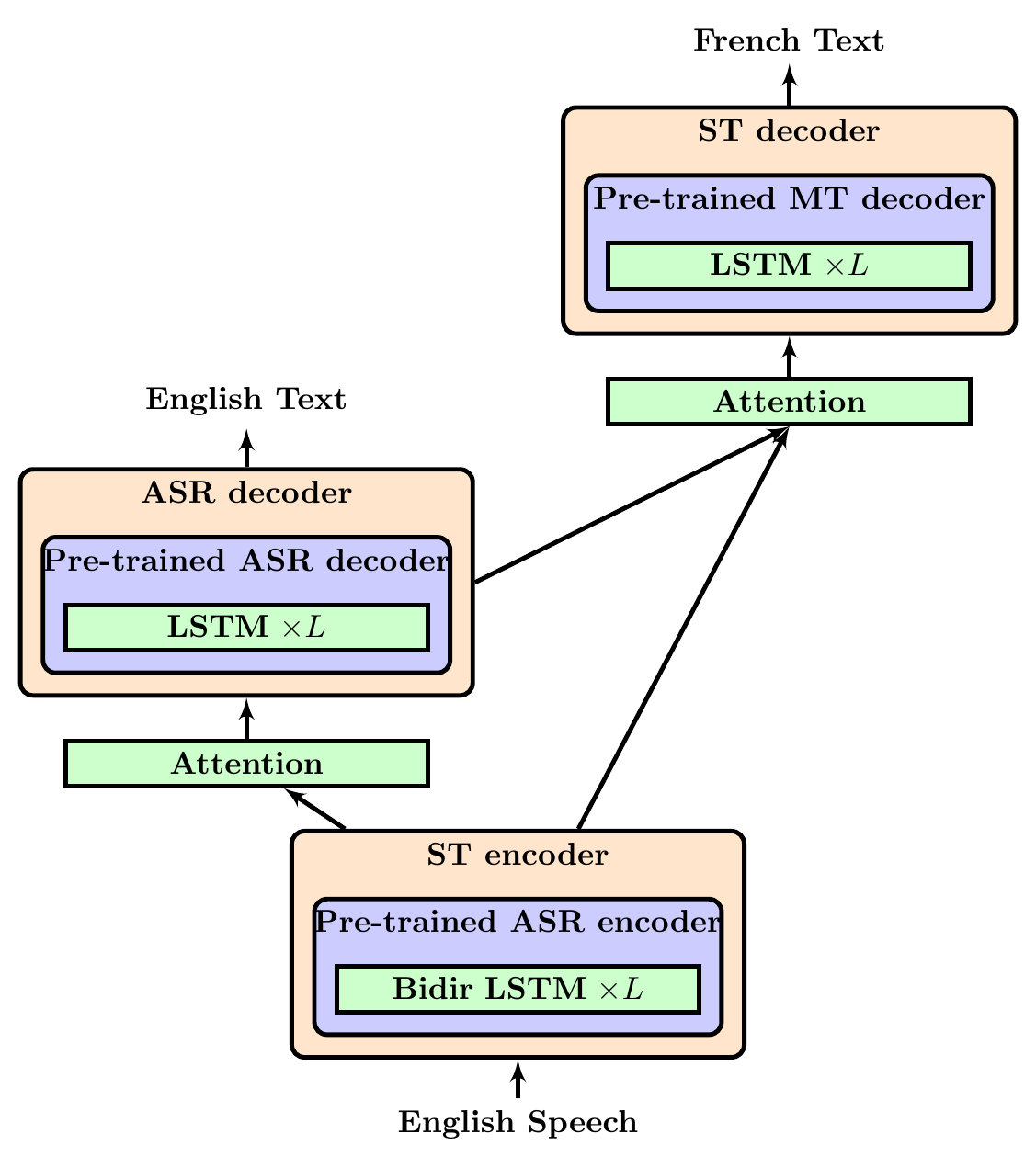}}
        \caption{multitask-tied triangle}
        \label{fig:multitask- tied triangle}
    \end{subfigure}
    \caption{Overview of end-to-end speech translation model architectures. Blue
blocks correspond to pre-trained components, and orange blocks are fine-tuned on the ST task. Green depicts the layers of architecture.}\label{fig:arch}
\end{figure*}

\section{Network Architectures}
\label{sec:network architecture}

%In speech translation, 
We assume a speech input observation of variable length $T$, $x_1^{T}$, a sequence of tokens of unknown length $J$ in the source language, $f_1^{J}$, and a sequence of target tokens of unknown length $I$, $e_1^{I}$. 
We define translation posterior probability as
$p(e_1^{I} | x_1^{T}) = \prod_{i=1}^{I} p(e_i| e_1^{i-1}, x_1^{T})$
where $T> I$, $J$. This can be modeled by either explicit use of the source sequence ($f_1^{J}$) as a pivot step or implicitly learn the representations.
The former can be a cascade of an ASR followed by an MT system and the latter is the usage of an end-to-end model. Here, we aim to address various kinds of end-to-end approaches.
% with or without the source sequence.
% however, in the multi-task learning, a sequence of transcriptions are used.

\subsection{Direct Model}
\label{sec:direct}
The direct model is the vanilla end-to-end network based on long short-term memories (LSTMs) \cite{hochreiter_97_LSTM} attention encoder-decoder architecture \cite{bahdanau_15_attention}. 
Here, we focus on LSTM-based models rather than the transformer \cite{vaswani_17_transformer, gangi-etal-2019-enhancing,st_kd_interspeech2019}.
The model is summarized in Equation \ref{eq:att}. 
A bidirectional LSTM (BLSTM) converts the input sequence into a sequence of encoder representations $h_1^{T}= h_1,\ldots, h_{T}$. 
% In order to handle the long speech utterances, we apply max-pooling in the time-dimension at multiple steps inside the speech encoder. 
% 
For speech encoders, to reduce the audio sequence length, we apply 3 max-pooling layers with a pool-size of 2 in the time-dimension at multiple steps between BLSTM layers.
For the input sequence $x_1^T$, we condense to $h_1^{T'}$, where $T' = T / 8$.
In the decoder, an LSTM generates an output sequence using an attention function outputting attention weights $\alpha_{i,t}$. 
The context vector $c_i$ is obtained as a weighted sum of encoder representations using these weights. $e_{i}$ is predicted by a linear transformation followed by a softmax layer, and the decoder state is updated to $s_i$ by a stack of LSTM layers. $L$ is the number of layers.
Figure \ref{fig:direct} illustrates an abstract overview of the direct model.

\begin{align}
&h_1^{T'} = (\operatorname{BLSTM}_{\rm{L}}
\circ \cdots
\circ \operatorname{max-pool}_1
\circ \operatorname{BLSTM}_1) (x_1^T) \nonumber \\
&\alpha_{i,t} = \operatorname{softmax} \big(\tanh (s_{i-1},h_{t}) \big) \text{,\hspace{0.5cm} } c_i = \sum_{t=1}^{T'} \alpha_{i,t} h_t  \nonumber \\
&p(e_{i}|e_1^{i-1},x_1^T) = \operatorname{softmax} \big( \operatorname{linear}(e_{i-1}, s_{i-1}, c_{i}) \big)  \nonumber\\
&s_i = \operatorname{LSTM}_{L}
\circ \cdots
\circ \operatorname{LSTM}_1 (e_{i}, s_{i-1}, c_{i}) \label{eq:att}
\end{align}

\subsection{Mutli-Task Model}
\label{sec:multitask}
As suggested in \cite{weiss_2017_directly}, in multi-task learning an auxiliary model is co-trained with the speech translation model by sharing some parameters. This auxiliary model can be an ASR or an MT model. We compare different multi-task training strategies as follows:
\\
\textbf{One-to-many}
As shown in Figure \ref{fig:multitask-one2many}, in the one-to-many method, an ASR model is used as the auxiliary co-trainer in which a speech encoder is shared between both tasks, while two independent decoders correspond to transcription and translation texts. 
The error is back-propagated via two decoders into the input and thus the final loss is computed as the weighted sum of the two losses.
Here, we choose $\lambda=0.5$.
\begin{align}
L = \lambda \log p_{s2s}(e_1^{I} | x_1^{T}) + (1-\lambda) \log p_{s2s}(f_1^{J} | x_1^{T})
\label{eq:multi:one2many}
\end{align}
\textbf{Many-to-one}
In many-to-one architecture, a text decoder is shared to generate a target translation by attending on two independent encoders, a speech encoder, and a text encoder. 
% A stack of BLSTM layers scans the audio features to compute a sequence of speech encoder states and another one is applied to encode the source sequence into $h^2_1,\cdots, h^2_{J}$.
Then the text decoder interchangeably attends on one of them.
Here, an MT model is co-trained along with the ST model (see Figure \ref{fig: multitask-many2one}).
% \begin{align}
% h^1_t = \text{LSTM} (x_{t}, h^1_{t-1}) \text{ \hspace{0.5cm}} h^2_j = \text{LSTM} (f_{j}, h^2_{j-1}) 
% \end{align}
\\
\textbf{Tied cascade}
Unlike the multitask network discussed above, where information shared either in the encoder or in the decoder, in the tied multi-task architectures \cite{anastasopoulos_2018_tied_multitask}, a higher-level intermediate
representation is provided.
In the tied cascade model, the decoder of the ST task is stacked on top of an ASR decoder and the ASR decoder is stacked on top of a speech encoder respectively (see Fig. \ref{fig:multitask- tied cascaded}).
Therefore, the second (ST) decoder attends only to the output states of the first (ASR) decoder. 
\\
\textbf{Tied triangle}
In the tied triangle model, the second decoder attends both on the encoder states of a speech encoder and on the states of the first text decoders as shown in Fig. \ref{fig:multitask- tied triangle}. 
In this architecture, two context vectors form the final representation.
% as inputs of the second decoder.
We use the greedy search to generate the first decoder's output and combine losses like Eq. \ref{eq:multi:one2many} in both tied cascade and triangle models.

\section{CTC Loss}
\label{sec:ctc}
CTC has been introduced in \cite{graves_2006_ctc} to solve the problem of unknown segmentation of the input sequence. It introduces a special blank state, which can appear at any time and represents that currently no token is recognized. Note that the blank state is not the same as whitespace, and is modeled as an own token.

A simple intuition is that the CTC module can not cooperate with the ST task since the translation task does not necessarily require the monotonic alignment of the speech and target sequences, unlike ASR.
But, based on the fact that an auxiliary CTC loss function has shown to help convergence in the training of ASR models \cite{zeyer_2018_att_asr}, we also apply it on top of the speech encoder only during training for the ST models. 
During training, given an input sequence $x_1^{T}$, the decoder predicts the frame-wise posterior distribution of $e_1^{I}$, whilst the CTC module predicts posterior distribution of $f_1^{J}$, referred to $p_{s2s}$ and $p_{ctc}$ respectively. We simply use the sum of their log-likelihood values as:
\begin{align}
L = \log p_{s2s}(e_1^{I} | x_1^{T}) +  \log p_{ctc}(f_1^{J} | x_1^{T})
\end{align}

We note that we also use CTC loss in the multi-task learning setups, however, one of the decoders predicts the frame-wise posterior distribution of the transcriptions. We believe CTC loss can help the ASR sub-task. Thus, Eq. \ref{eq:multi:one2many} is written as:
\begin{align}
L &= \lambda \log p_{s2s}(e_1^{I} | x_1^{T}) \nonumber \\
 &+ (1-\lambda) \Big(\log p_{s2s}(f_1^{J} | x_1^{T}) + \log p_{ctc}(f_1^{J} | x_1^{T}) \Big)
\end{align}

\section{Pre-training}
\label{sec:pretraining}

Pre-training greatly improves the performance of end-to-end ST networks. 
In this context, pre-training refers to pre-train the model on a high-resource task, and then fine-tune its parameters on the ST data. 
In Figure \ref{fig:arch}, the blue blocks refer to the pre-trained components and the orange blocks mean fine-tuning on the ST data. 
Pre-training can be done in different ways as proposed in the literature. The common way is to use an ASR encoder and an MT decoder to initialize the parameters of the ST network correspondingly \cite{bansal_2018_low_resource}. Surprisingly, using an ASR model to pre-train both the encoder and the decoder of the ST model works well \cite{bansal_2019_pretraining_asr}. 

\begin{table}[h]
\begin{center}
\caption{Training data statistics.}
\scalebox{0.9}{%
\label{tab:stat}
\begin{tabular}{lll|ll}
\hline
\multirow{2}{*}{\bfseries Task} & \multicolumn{2}{c}{\bfseries IWSLT En$\to$De} & \multicolumn{2}{c}{\bfseries LibriSpeech En$\to$Fr} \\ \cline{2-5}
     & \# of seq.  & hours    & \# of seq.    & hours    \\ \hline
ASR  &  92.9k    &  207h &  61.3k  &  130h       \\
ST   &  171.1k   &  272h &  94.5k  &  200h        \\
MT   &  32M      &   -   &  94.5k   &  - \\ \hline        
      
\end{tabular}
}
\end{center}
\end{table}

\section{Experiments}
\label{sec:expriments}
\subsection{Datasets and Metrics} 
\label{sec:dataset}

We have done our experiments on two ST tasks: the IWSLT TED En$\to$De \cite{Cho_2014_ted,Niehues_2018_iwslt2018}\footnote{https://sites.google.com/site/iwsltevaluation2018/Lectures-task} and the LibriSpeech En$\to$Fr \cite{Kocabiyikoglu_2018_librispeech, berard_2018_librispeech}\footnote{https://persyval-platform.univ-grenoble-alpes.fr/DS91/detaildataset}. Table \ref{tab:stat} shows the training data statistics.
\\
\textbf{IWSLT En$\to$De:}
We use the TED-LIUM corpus (excluding the black-listed talks) including 207h and the IWSLT speech translation TED corpus with 272h of speech data, in total 390h. 
80-dimensional Mel-frequency cepstral coefficients (MFCC) features are extracted. 
Similar to \cite{apptek_2018_st}, we automatically recompute the provided audio-to-source-sentence alignments to reduce the problem of speech segments without a translation.
We randomly select a part of our segments as our cross-validation set and choose dev2010 and test2015 as our development and test sets with 888 and 1080 segments respectively. We select our checkpoints based on the dev set.
For the MT training, we use the TED, OpenSubtitles2018, Europarl, ParaCrawl, CommonCrawl, News Commentary, and Rapid corpora resulting in 32M sentence pairs after filtering noisy samples.
\\
\textbf{LibriSpeech En$\to$Fr:}
Similar to \cite{berard_2018_librispeech}, to increase the training data size, we add the original translation and the Google Translate reference provided in the dataset package. It results in 200h of speech corresponding to 94.5k segments for the ST task.
We extract 40-dimensional Gammatone features \cite{schluter_2007_gammatone} using
the RASR \cite{wiesler_2014_rasr}.
% For the MT training, besides the LibriSpeech corpus, we utilize the OpenSubtitles2018 and the Europarl corpora resulting in 43M sentence pairs, however, later we observed that they are out-of-domain and barely helped the performance.
For MT training, we utilize no extra data.
The dev and test sets contain 2h and 4h of speech, 1071 and 2048 segments respectively. The dev set is used as our cross-validation set and checkpoint selection.

For the IWSLT, in addition to the ST data, we benefit from weakly supervised data (paired ASR and MT data) both in multi-task learning and in pre-training. Whereas for the LibriSpeech, we only use the ST data in the multi-task scenario to see whether any gain comes from the data or the model itself.

For both tasks, we remove the punctuation only from the transcriptions (i.e. the English text) and keep the punctuation on the target side. We note that by doing so we tend to explore whether the MT models can implicitly capture the punctuation information without any additional component. Furthermore, we do not need to automatically enrich the ASR's output with punctuation marks for the cascade pipeline. \texttt{Moses} toolkit \cite{koehn_07_moses}\footnote{http://www.statmt.org/moses/?n=Moses.SupportTools}  is used for tokenization. We employ frequent casing for the IWSLT tasks while lowercase for the LibriSpeech. There, the evaluation of the IWSLT En$\to$De is case-sensitive, while that of the LibriSpeech is case-insensitive\footnote{We do the case-insensitive evaluation to be comparable with the other works, however, it is not clear which \BLEU script they used \cite{chung_2019_unsupervised_st, berard_2018_librispeech}.}.
The translation models are evaluated using the official scripts of WMT campaign, i.e. \BLEU~\cite{papineni_02_bleu} computed by \texttt{mteval-v13a}\footnote{ftp://jaguar.ncsl.nist.gov/mt/resources/mteval-v13a.pl} and \TER~\cite{snover_06_ter} computed by \texttt{tercom}\footnote{http://www.cs.umd.edu/~snover/tercom/}. \WER~is computed by \texttt{sclite}\footnote{http://www1.icsi.berkeley.edu/Speech/docs/sctk-1.2/sclite.htm}.

\begin{table}
\begin{center}
\caption{ASR results measured in \WER~[\%].}
\label{tab:asr_results}
\begin{tabular}{lrr}

\hline
\multirow{2}{*}{\bfseries Task} & \multicolumn{2}{c}{\bfseries \WER[$\downarrow$]}  \\
& \bfseries  dev  &  \bfseries test \\ \hline
IWSLT En$\to$De       &  12.36  &   13.80   \\ 
LibriSpeech En$\to$Fr &  \phantom{0} 6.47 & \phantom{0} 6.47  \\ \hline

\hline
\end{tabular}
\end{center}
\end{table}

\subsection{Models} 
\label{sec:models}

In our experiments, we build an ASR, an MT, and different ST models. The ASR and MT models are used for building the cascade pipeline as well as pre-training the ST networks. All models are based on the attention model described in Section \ref{sec:network architecture} with one or two encoder(s) or decoder(s) according to the architecture.

For both tasks, byte pair encoding (BPE) \cite{sennrich_16_bpe} with $20$k merge operations on the MT data (both source and target side), and $10$k symbols on the ASR transcriptions is used.
We map all tokens into embedding vectors of size 620. Both speech and text encoders are built with 6 stacked BLSTM layers equipped with 1024 hidden size. 
Similar to \cite{zeyer_2018_att_asr}, we apply layer-wise pre-training, where we start with two encoder layers. 
As explained before, for speech encoder, we utilize max-pooling between BLSTM layers to reduce the audio sequence length. 
The decoders are a 1-layer unidirectional LSTM of size 1024.
Our attention component is composed of a single head additive attention with alignment feedback \cite{tu2016ACL,bahar_2017_rwth}.
To enable pre-training, for ST models, we use the same architecture to the ASR encoder and the same architecture to the MT decoder.

We train models using Adam update rule \cite{kingma_14_adam} with a learning rate between 0.0008 to 0.0001. We apply dropout of 0.3 \cite{srivastavad_14_dropout}, and label smoothing \cite{pereyra_2017_label_smoothing} with a ratio of 0.1.
We lower the learning rate with a decay factor of 0.9 and wait for 6 consecutive checkpoints. Maximum sequence length is set to 75 sub-words. All batch sizes are specified to be as big as possible to fit in a single GPU memory. 
A beam size of 12 is used during the search.
The models are built using our in-house implementation of end-to-end models in
\texttt{RETURNN} \cite{zeyer_18_returnn}
that relies on \texttt{TensorFlow} \cite{tensorflow}.
The code and the configurations of the setups are available online~\footnote{https://github.com/rwth-i6/returnn}~\footnote{https://github.com/rwth-i6/returnn-experiments}.

\begin{table}
\begin{center}
\caption{MT results measured in \BLEU~[\%] and \TER~[\%] using ground truth source text.}
\label{tab:mt_results}
\begin{tabular}{lllll}

\hline
\multirow{2}{*}{\bfseries Task} & \multicolumn{2}{c}{\bfseries \BLEU[$\uparrow$]} & \multicolumn{2}{c}{\bfseries \TER[$\downarrow$]}  \\
& \bfseries  dev  & \bfseries  test &  \bfseries dev  &  \bfseries test\\ \hline
IWSLT En$\to$De       &  30.50 & 31.50 &  50.57  & -  \\ 
LibriSpeech En$\to$Fr &  20.11 & 18.22 &  65.27  & 67.71 \\ \hline

\hline
\end{tabular}
\end{center}
\end{table}

\begin{table}
\begin{center}
\caption{ST results using cascaded pipeline of end-to-end ASR and MT measured in \BLEU~[\%] and \TER~[\%].}
\label{tab:st_results}
\begin{tabular}{lllll}
\hline
\multirow{2}{*}{\bfseries Task} & \multicolumn{2}{c}{\bfseries \BLEU[$\uparrow$]} & \multicolumn{2}{c}{\bfseries \TER[$\downarrow$]}  \\
& \bfseries  dev  & \bfseries  test &  \bfseries dev  &  \bfseries test\\ \hline
IWSLT En$\to$De       &  24.67 & 24.43 & 58.86  &  62.52 \\ 
LibriSpeech En$\to$Fr &  17.31 & 15.74 & 69.08   & 70.59 \\ \hline

\hline
\end{tabular}
\end{center}
\end{table}

\begin{table*}
\begin{center}
\caption{ST results for different architectures measured in \BLEU~[\%] and \TER~[\%].}
\scalebox{0.9}{%
\label{tab:archs}
\begin{tabular}{lllllllll}

\toprule

\multirow{2}{*}{\bfseries Method} & \multicolumn{4}{c}{\bfseries En$\to$De} & \multicolumn{4}{c}{\bfseries En$\to$Fr} \\
 & \multicolumn{2}{c}{\bfseries dev} & \multicolumn{2}{c}{\bfseries test} & \multicolumn{2}{c}{\bfseries dev} & \multicolumn{2}{c}{\bfseries test}\\
& \bfseries \BLEU         & \bfseries \TER        & \bfseries \BLEU         & \bfseries \TER    & \bfseries \BLEU         & \bfseries \TER        & \bfseries \BLEU         & \bfseries \TER    \\ 
\midrule
direct            & 14.80 & 69.81 & 14.86 & 72.49 & 15.71 & 75.86 & 14.69 & 76.53 \\
\quad + CTC    & 17.86 & 66.32 & 16.50 & 70.40 & \textbf{16.41} & \textbf{74.17} & \textbf{15.11} & \textbf{75.76} \\ 
\midrule

one-to-many   & 17.25 & 66.98 & 16.29 & 71.05 & 15.31 & 74.79 & 14.28 & 76.84 \\
\quad + CTC & 17.77 & \textbf{66.15} & 16.94 & 70.04 & 16.01 & 73.93 & 14.46 & 76.25\\ 
\midrule

many-to-one    & \textbf{18.17} & 67.23 & \textbf{18.06} & \textbf{69.83} & 11.88 & 79.69 & 11.45 & 81.69 \\
\quad + CTC    & 17.54 & 65.47 & 16.96 & 71.20 & 11.76 & 79.51 & 11.52 & 80.95 \\ \midrule

tied cascaded  & 15.36 & 69.91 & 14.85 & 73.21 & 13.28 & 77.08 & 12.46 & 79.14 \\
\quad + CTC    & 16.60 & 68.73 & 15.68 & 72.44 & 13.75 & 76.88 & 12.83 & 79.15 \\ \midrule

tied triangle    & 14.24 & 70.83 & 13.83 & 74.30 & 12.76 & 78.97 & 12.09 & 79.43 \\
\quad + CTC    & 15.94 & 68.50 & 14.96 & 71.83 & 13.39 & 77.24 & 12.68 & 78.87 \\ 
\bottomrule
\end{tabular}
}
\end{center}
\end{table*}

\section{Results}

Table \ref{tab:asr_results} and \ref{tab:mt_results} show the ASR and MT results on IWSLT and LibriSpeech tasks respectively. On the test sets, we gain 13.80\% and 6.47\% \WER. 
% The MT task on LibriSpeech seems more challenging as both scores are lower. 
We obtain 31.50\% \BLEU on the IWSLT and respectively 18.22\% \BLEU on the LibriSpeech by pure MT.
Table \ref{tab:st_results} also shows our baseline that is the traditional cascade pipeline where the output of our ASR model, a sequence of tokens, is fed as the input to our MT system. We achieve 24.43\% \BLEU and 62.52\% \TER on the IWSLT test set and 15.74\% \BLEU and 70.59\% \TER on the LibriSpeech. As expected, the ST systems are behind the pure MT models trained using ground truth source text (cf. \ref{tab:mt_results} and \ref{tab:st_results}).

\subsection{Network Architectures}
The results of five different end-to-end ST architectures are listed in Table \ref{tab:archs}.
On the IWSLT task, for one-to-many and many-to-one approaches we utilize additional ASR and MT data for the respective auxiliary tasks. Therefore, we observe improvements in the performance of both compared to the direct model. The one-to-many method improves by an average of 1.98\% \BLEU and 2.13\% \TER and the many-to-one approach by an average of 3.29\% \BLEU and 2.61\% \TER. \\
On LibriSpeech, we only work with the ST data and notice no real improvement over the direct model. It supports the intuition that the gain in the multi-task training relies on more data rather than better learning, however, it does not hurt. \\
Both the tied cascade and triangle models perform worse than the direct models. We also notice that the cascade models generally seem to perform slightly better than the triangle models (see Table \ref{tab:archs}).
Our results show that the auxiliary data in different multi-task learning is beneficial only up to some degree and that the multi-task models are significantly dependent on the end-to-end data. Our justification is that the multi-task training that utilizes auxiliary models on additional data, discards many of the additionally learned parameters, hence it does not lead to better results than the direct model at the end.
We also think a stack of 2 decoders in the tied methods makes the optimization harder. That's why their performance is behind the other architectures.
\subsection{CTC Loss}

Similar to \cite{zeyer_2018_att_asr}, we use CTC as an additional loss mainly to help the initial convergence, also as a mean for regularization. 
The results are listed in Table \ref{tab:archs} for each of the discussed architectures. Interestingly, due to better optimization, in almost all cases, the CTC loss helps the performance in terms of both \BLEU and \TER. As shown, the largest improvement is on the direct model with 1.64\% in \BLEU and 2.09\% in \TER on the test set and 3.06\% in \BLEU and 3.50\% in \TER on the dev set for the IWSLT task. The same trend can be seen on the LibriSpeech with a smaller impact. We believe that, the easier the optimization is, the more beneficial the CTC loss is. That's why it helps the direct model more than the other architectures. \\
% We have also noticed that CTC in combination with pre-training barely helps. 
It is important to highlight that in principle, the CTC loss is similar to adding a second decoder to predict the transcription equipped with an empty token. On the many-to-one model, CTC loss leads to degradation in performance. Our interpretation is that since we have both speech and text encoders and we only equip the speech encoder with the CTC, it might hurt the learning phase of the text encoder. Because the parameters are updated independently for each encoder.

\begin{table}[t]
\begin{center}
\caption{Different pre-training scheme for En$\to$De corresponding to the blue blocks of Fig. \ref{fig:arch}. \lq\lq-\rq\rq: failed results. $^1$: similar to \cite{bansal_2019_pretraining_asr}, we pre-train German ST/MT decoder using a pre-trained English ASR decoder.}
\scalebox{0.72}{%
\label{tab:pre-training}
\begin{tabular}{lllll}
\toprule

\multirow{2}{*}{\bfseries Method} & \multicolumn{2}{c}{\bfseries dev} & \multicolumn{2}{c}{\bfseries test} \\
 & \bfseries \BLEU & \bfseries \TER        & \bfseries \BLEU         & \bfseries \TER        \\ \hline
\bfseries direct \\
 ASR enc.               & 20.15 & 64.29 & 19.41 & 67.52 \\
 MT dec.                & - & - & - & -  \\
 ASR enc.+MT dec. (w ASR dec)$^1$  & 19.92 & 62.48 & 19.56 & 66.68 \\
% \quad ASR enc.+MT dec.     & 19.52 & 63.16 & 19.26 & 68.4 \\
 ASR enc.+MT dec.     & -& - & - & - \\
\quad+ adopter        & \textbf{21.07} & 62.14 & \textbf{20.74} & \textbf{65.45} \\ 
\midrule

\bfseries one-to-many   \\
MT dec.               & - & - & - & -  \\
ASR enc.+ASR dec      & 18.89 & 64.39 & 17.68 & 68.73 \\
ASR enc.+ASR dec.+MT dec. (w ASR dec)$^1$ & 18.69 & 64.31 & 17.27 & 68.75 \\
ASR enc.+ASR dec.+MT dec.  & 19.39 & 63.92 & 18.07 & 68.35 \\
\quad + adopter        & \textbf{21.07} & \textbf{61.80} & 19.17 & 67.86 \\ 
\midrule

\bfseries many-to-one \\
ASR enc.              & 17.84 & 65.95 & 16.75 & 70.65 \\
MT dec.               & - & - & - & - \\
ASR enc.+MT enc.+MT dec. (w ASR dec)$^1$  & 18.95 & 64.78 & 18.23 & 68.93 \\
ASR enc.+MT enc.+MT dec. & 18.13 & 65.78 & 17.10 & 70.33 \\
\quad+ adopter        & \textbf{21.17} & 62.12 & 19.77 & 66.79 \\ 
\midrule
\bfseries tied cascade  \\
MT dec.             & - & - & - & - \\
ASR enc.+ASR dec.  & 16.44 & 73.13 & 15.45 & 78.03 \\
ASR enc.+ASR dec.+MT dec. & 14.98 & 75.55 & 14.56 & 79.73  \\
\quad+ adopter       & 15.91 & 73.50 & 14.19 & 83.62 \\
\midrule
\bfseries tied triangle \\
MT dec.             & - & - & - & - \\
ASR enc.+ASR dec. & 16.67 & 67.96 & 15.59 & 71.88 \\
ASR enc.+ASR dec.+MT dec. & 16.49 & 67.14 & 15.37 & 71.65 \\
\quad+ adopter           & 18.52 & 64.38 & 18.20 & 67.98 \\ 
\bottomrule

\end{tabular}
}
\end{center}
\end{table}

\subsection{Pre-training}
We have also done several experiments to figure out the optimal way of utilizing pre-training for the end-to-end ST models. In theory, pre-training should not hurt but in practice, it often does because the pre-trained components are not incorporated properly together. In this case, it might lead to an optimization problem in which the training is not robust to new training data variations. We carry out different schemes in various architectures. The results are shown in Table \ref{tab:pre-training}. The first column of the table corresponds to blue blocks in Figure \ref{fig:arch}. For instance, \texttt{ASR enc+MT dec} refers to the case in which we initialize the network's components using a pre-trained ASR encoder and MT decoder. 
% In the figure, the blue boxes are related to pre-training different components

In the first attempt, 
we have tried to only initialize the speech encoder using a pre-trained ASR encoder. As it can be seen in the table, it gives a boost in terms of both \BLEU and \TER for the direct but not for the many-to-one model where it leads to worse performance. 
%It helps the direct model the most. 
Our explanation is that, for multi-task models, we have an extra component that has to be trained jointly and it makes the optimization task harder, even with better initialization using pre-trained models. We should note that we have not tried using \texttt{ASR enc} pre-training for the other models  since we can initialize both the encoder and the first decoder using an entire ASR model, \texttt{ASR enc+ASR dec} (discussed later).
Then, we take an MT decoder to initialize the text decoder.
We observe that pre-training the text decoder seems to harm the models in all cases such that no reasonable performance is obtained (shown by \lq\lq-\rq\rq). Our observations indicate that the pre-trained MT decoder expects to be jointly trained with a text encoder, not a speech encoder, as it has been trained with a text encoder beforehand.  
Therefore, one should integrate these two components in a better way. 

This huge degradation has led to further investigations where we study why pre-training of text decoder using an MT model hurts.
To explain this justification, we first try pre-training both the encoder and the decoder using our ASR model as suggested in \cite{bansal_2019_pretraining_asr}. Since the ASR decoder is already familiar with the ASR encoder, this problem should be disappeared. As shown for the direct model, it performs comparable with the case where we only used ASR encoder. We also apply the method for the other multi-task models with ASR decoders. As listed in Table \ref{tab:pre-training}, in almost all cases, it greatly outperforms the non-pre-trained baseline model. 
% For the tied cascade and triangle, \lq\lq ASR enc.+ASR dec.\rq\rq helps

In the next step, we pre-train both the speech encoder and the text decoder(s) at the same time. Again as it has been shown, for the direct model, coupling the pre-trained ASR encoder and MT decoder fails training, however, in the other architectures, this setup outperforms the baseline. We have also observed that if we pre-train all layers except the attention component, \texttt{ASR enc+ASR MT} works well for the direct model.
% Only for tied cascade and triangle approaches we have no performance. We believe that this is due to the fact that the second decoder sees the output of the ASR decoder as the input. However it expects to see the output of an ASR encoder.
Based on these observations we add an additional layer (one BLSTM) as an \texttt{adapter} component to familiarize
the input of the pre-trained decoder with the output of
the pre-trained encoder. 
We insert it on the top of the component where we want to be coupled smoothly. It means, for direct, one-to-many and many-to-one methods, we add it on the top of the pre-trained speech encoder. For the tied cascade and triangle, we add it on the top of the ASR pre-trained decoder. 
We train the adaptor component jointly without freezing the parameters. 
This helps the fine-tuning stage on the ST data by 6.08\% in \BLEU and 7.36\% in \TER on average for the direct model at most. In general, we observe better results for all architectures except for the tied cascade system.
Interestingly the adopter approach works better for the triangle architecture compared to the tied cascade. Possibly here the decoder can utilize the adopter layers better to focus more and connect to the output of the ASR encoder rather than the decoder, mimics similar to the direct approach and provides better results.

\begin{table}
\begin{center}
\caption{Comparison of test sets with the literature. $^1$: the evaluation is without punctuation. To be comparable with other works, we note that the LibriSpeech results in this table are case-insensitive \BLEU computed using \texttt{multi-bleu.pl} script \cite{koehn_07_moses}.}
\scalebox{0.90}{%
\label{tab:others}
\begin{tabular}{lcc}

\toprule
\multirow{2}{*}{\bfseries Method} & \multicolumn{1}{c}{\bfseries En$\to$De} & \multicolumn{1}{c}{\bfseries En$\to$Fr} \\ 
&\bfseries test  & \bfseries test \\ 
\midrule
\bfseries other works  \\ 
\quad direct \cite{liu_2018_iwslt} & 20.07 & - \\
\quad cascade pipeline \cite{liu_2018_iwslt}  & 25.99 & - \\
\quad direct \cite{berard_2018_librispeech}         & - & 13.30\\
\quad multi-task  \cite{berard_2018_librispeech}     & - & 13.40\\
% \quad cascade pipeline  \cite{berard_2018_librispeech}  & -  & 14.60\\
\quad unsupervised$^1$   \cite{chung_2019_unsupervised_st}   & - &12.20  \\ 
\quad transformer \cite{gangi-etal-2019-enhancing}   & - & 13.80  \\
% \quad direct transformer   \cite{st_kd_interspeech2019}   & - & 13.15  \\ 
\quad transformer+pretraining   \cite{st_kd_interspeech2019}   & - & 14.30  \\ 
\quad\quad + knowledge distillation   \cite{st_kd_interspeech2019}   & - & 17.02  \\
% \quad + knowledge distillation   \cite{st_kd_interspeech2019}   & - & 17.02  \\ 

\toprule
\bfseries this work  \\  
% \quad cascade pipeline  & 24.43  & 15.74\\
\quad direct+pretraining+adaptor   & 20.74& 16.80 \\ 

\bottomrule
\end{tabular}
}
\end{center}
\end{table}

Finally, we also compare our models with the literature in Table \ref{tab:others}. We use our best setup which is a direct model using an adapter on top of pre-training for the comparison in this table. On the IWSLT test set, our model outperforms the winner of the IWSLT2018 evaluation campaign by 0.67\% in \BLEU. On the LibriSpeech test set, our model outperforms both the LSTM-based and the transformer models and slightly behind the knowledge distillation method.

\section{Conclusion}
\label{sec:conclusion}

In this work, we have studied the performance of various architectures for end-to-end speech translation, where a moderate amount of speech translation data, as well as weakly supervised data, i.e. ASR or MT pairs, are available.
We have demonstrated the effect of pre-training of the network's components and explored how to efficiently couple the encoder and the decoder by adding an additional layer in between. This extra layer allows for better joint learning and gives performance boosts. 
Moreover, CTC loss can be an important factor for the end-to-end ST training since it leads to better performance as well as faster convergence. 
% It gives an average boost of 2.4\% and 0.5\% in \BLEU on En$\to$De and En$\to$Fr respectively. Therefore, it is an appropriate choice for direct modeling. 
% Finally, we showed that our direct model facilitated with a proper pre-training scheme works reasonably good such that it outperforms the cascade system on the LibriSpeech task.

\section{Acknowledgements}
\label{sec:acknowledgements}
\begin{wrapfigure}{l}{0.15\textwidth}
\vspace{-4mm}
    \begin{center}
        \includegraphics[width=0.17\textwidth]{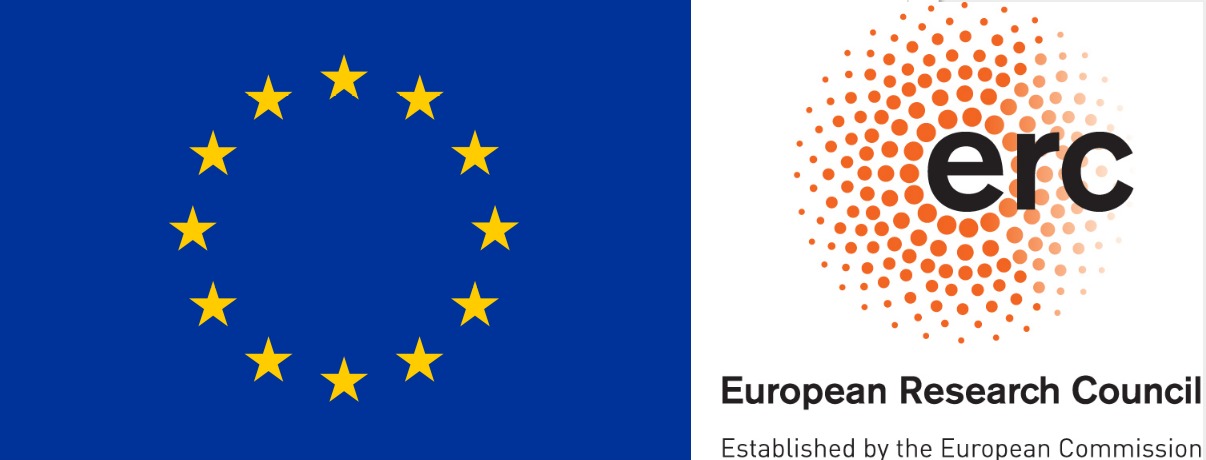} \\
       \vspace{1mm} 
        \includegraphics[width=0.17\textwidth]{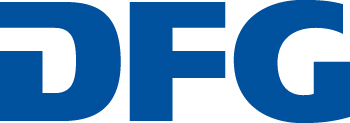}
    \end{center}
\vspace{-4mm}
\end{wrapfigure}
This work has received funding from the European Research Council (ERC) under the European Union's Horizon 2020 research and innovation programme (grant agreement No 694537, project "SEQCLAS"), the Deutsche Forschungsgemeinschaft (DFG; grant agreement NE 572/8-1, project "CoreTec") and from a Google Focused Award. The work reflects only the authors' views and none of the funding parties is responsible for any use that may be made of the information it contains. 
% The GPU cluster used for the experiments was partially funded by Deutsche Forschungsgemeinschaft (DFG) Grant INST 222/1168-1.


\bibliographystyle{IEEEbib}

% Give us some more space:
% Tune this later if needed, or just uncomment if not needed.

% This works perfect! :)
%\setstretch{0.95}

%\def\baselinestretch{0.8}
%\let\normalsize\small\normalsize
%\SetTracking{encoding=*}{-15}\lsstyle  % still somewhat ok
%\SetTracking{encoding=*}{-85}\lsstyle

%\renewcommand{\baselinestretch}{0.1}\normalsize
% http://tex.stackexchange.com/questions/93859/condense-the-space-between-bibliographic-entries
\let\OLDthebibliography\thebibliography
\renewcommand\thebibliography[1]{
  \OLDthebibliography{#1}
  \setlength{\parskip}{-2pt}
  \setlength{\itemsep}{0pt plus 0.07ex}
}

\bibliography{Template_Regular}

\end{document}